\def\year{2022}\relax
\documentclass[letterpaper]{article} 
\usepackage{aaai22}  
\usepackage{times}  
\usepackage{helvet}  
\usepackage{courier}  
\usepackage[hyphens]{url}  
\usepackage{graphicx} 
\usepackage{natbib}  
\usepackage{caption} 
\usepackage{bbding} 
\DeclareCaptionStyle{ruled}{labelfont=normalfont,labelsep=colon,strut=off} 
\usepackage{algorithm}
\usepackage{algorithmic}
\usepackage{multirow}
\usepackage{booktabs}
\usepackage{diagbox}
\usepackage[switch]{lineno}
\usepackage[colorlinks,
            linkcolor=black,
            anchorcolor=black,
            citecolor=black]{hyperref}
\usepackage[symbol]{footmisc}

\usepackage{newfloat}
\usepackage{listings}
\floatstyle{ruled}
\newfloat{listing}{tb}{lst}{}
\floatname{listing}{Listing}

\title{Social Interpretable Tree for Pedestrian Trajectory Prediction}
\author {
    Liushuai Shi\textsuperscript{\rm 1},
    Le Wang\textsuperscript{\rm 2}\footnote{Corresponding author.},
    Chengjiang Long\textsuperscript{\rm 3} \\
    Sanping Zhou\textsuperscript{\rm 2},
    Fang Zheng\textsuperscript{\rm 1},
    Nanning Zheng\textsuperscript{\rm 2},
    Gang Hua\textsuperscript{\rm 4}
}
\affiliations {
    \textsuperscript{\rm 1}School of Software Engineering, Xi'an Jiaotong University\\
    \textsuperscript{\rm 2}Institute of Artificial Intelligence and Robotics, Xi'an Jiaotong University\\
    \textsuperscript{\rm 3}JD Finance America Corporation\\
    \textsuperscript{\rm 4}Wormpex AI Research\\
    \{shiliushuai,zhengfang\}@stu.xjtu.edu.cn, \{lewang,spzhou,nnzheng\}@xjtu.edu.cn, \{cjfykx, ganghua\}@gmail.com
}

\usepackage{bibentry}
\usepackage{amsmath}
\usepackage{amssymb}
\begin{document}

\maketitle
\begin{abstract}

Understanding the multiple socially-acceptable future behaviors is an essential task for many vision applications. 
In this paper, we propose a tree-based method, termed as Social Interpretable Tree~(SIT), to address this multi-modal prediction task, where a hand-crafted tree is built depending on the prior information of observed trajectory to model multiple future trajectories. 
Specifically, a path in the tree from the root to leaf represents an individual possible future trajectory. 
SIT employs a coarse-to-fine optimization strategy, in which the tree is first built by high-order velocity to balance the complexity and coverage of the tree and then optimized greedily to encourage multimodality.
Finally, a teacher-forcing refining operation is used to predict the final fine trajectory.
Compared with prior methods which leverage implicit latent variables to represent possible future trajectories, the path in the tree can explicitly explain the rough moving behaviors~(\emph{e.g.}, go straight and then turn right), and thus provides better interpretability.
Despite the hand-crafted tree, the experimental results on ETH-UCY and Stanford Drone datasets demonstrate that our method is capable of matching or exceeding the performance of state-of-the-art methods.
Interestingly, the experiments show that the raw built tree without training outperforms many prior deep neural network based approaches.
Meanwhile, our method presents sufficient flexibility in long-term prediction and different best-of-$K$ predictions.
\textit{Code: \url{https://github.com/lssiair/SIT}}
\end{abstract}

\section{Introduction}
Pedestrian trajectory prediction plays an essential role in many vision systems, \emph{e.g.}, the automatic vehicle understands the future trajectory of the pedestrian to prevent the accident, and the monitoring system recognize the abnormal action in advance by predicting the future trajectory of human.

\begin{figure}[t]
\centering
\includegraphics[width=0.99\columnwidth]{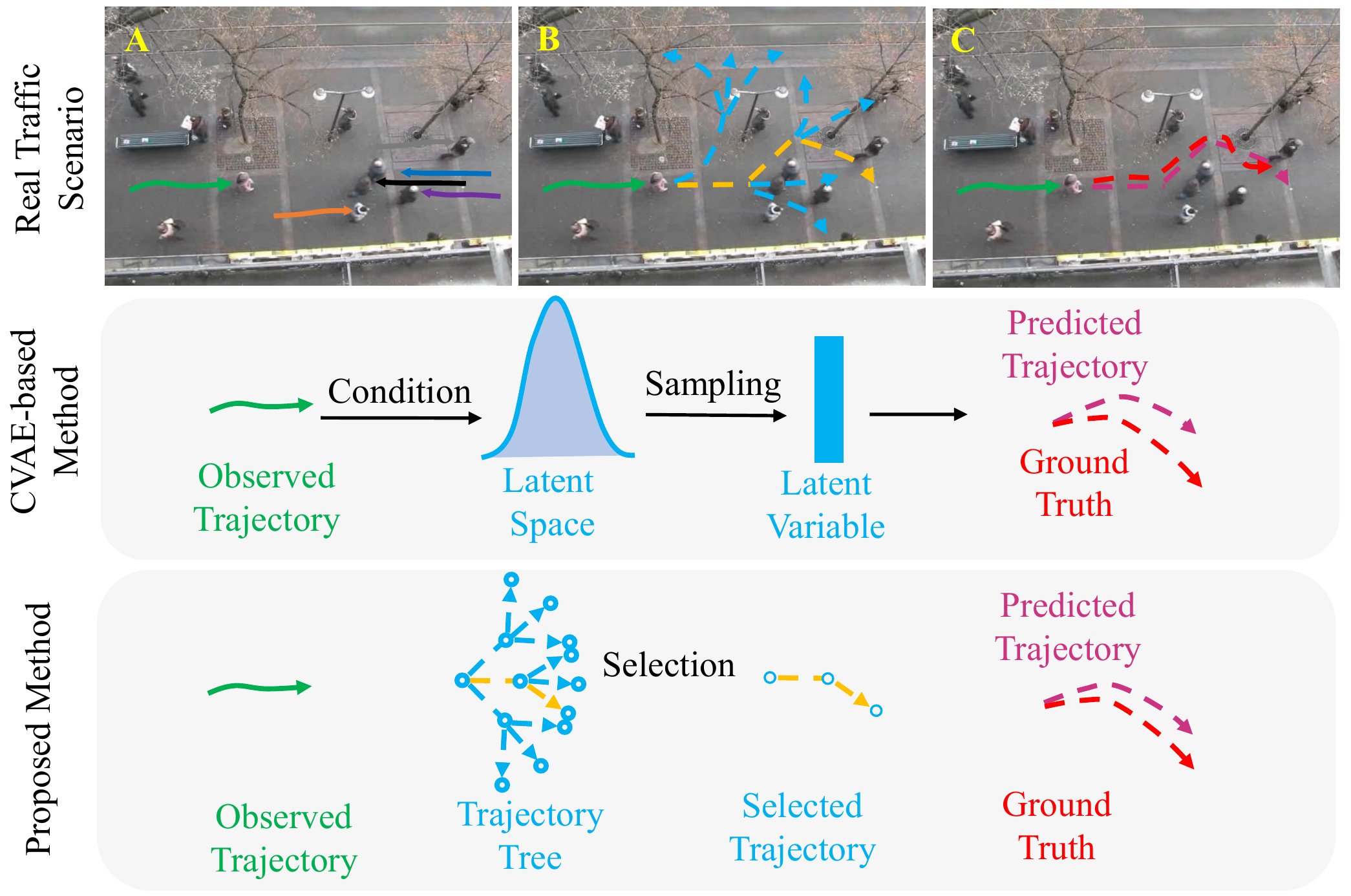} 
\caption{Illustration of real traffic scenario and comparison of our method with CVAE-based method. The first row illustrates the future trajectory is largely uncertain only referencing observed trajectory~(A). The future trajectory is multi-modal and thus can be represented by a tree~(B). The yellow closest path is refined to predict future trajectory~(C).
The second and third rows describe the process of CVAE-based method and our proposed method, respectively.
}
\label{fig1}
\vspace{-0.5cm}
\end{figure}

In a real traffic scenario illustrated in Figure~\ref{fig1}~(first row), due to the intrinsic randomness of pedestrians' moving and intangible various intent only based on observed trajectory~(A), the future trajectory is largely uncertain and naturally multi-modal, which means there are multiple possible trajectories that pedestrian could take~(B).
One kind of approaches~\cite{sgan, pec} to model this multi-modal future trajectory embed them into an implicit latent space~(second row) generated by the conditional variational autoencoder~(CVAE) or generative adversarial network~(GAN).
Then, multiple latent variables sampled repeatedly from the generated latent space are used to represent multimodality.
Despite the significant performance, the latent variables still suffer from uninterpretability, and such models~(CVAE, GAN) are persecuted by the problem of model collapse~\cite{arjovsky2017towards}. 
What's more, the sampling operation could result in performance variance due to the disturbance from randomness.

To cope with those problems, we propose to model the multi-modal future trajectory into a tree as shown in Figure~\ref{fig1}~(third row), where the paths from the root to leaf in the tree could represent multimodality naturally and the closest path~(yellow path) with ground truth is selected to obtain the final fine-grained predicted trajectory.
Compared with the latent variable, the path in the tree could explain rough movement behaviors, \emph{e.g.}, the yellow path expresses go straight and then turn right, and thus can provide well interpretability.
Furthermore, the sampling operation is replaced with the ``selection" to ensure obtaining stable results.

Inspired by the interpretable tree, we propose the Social Interpretable Tree~(SIT) to predict multi-modal future trajectories.
SIT first builds a future trajectory tree to generate plausible future trajectories according to the velocity of the observed trajectory.
To obtain concise representation, the tree is specified with a ternary tree, in which the tree splits in three directions, \emph{i.e.}, go straight, turn left and right with a specific angle, at each time step.
The turn round and keep still can be viewed in the specific cases of go straight and turn left or right, respectively.
Since the complexity of the tree grows exponentially as the depth increases, we propose to build a coarse trajectory tree~(CTT) to balance the complexity and coverage of the tree.
Instead of splitting time step by time step, the CTT splits through multi-time steps recursively and the high-order velocity in this temporal interval of observed trajectory is considered as the split direction.   

After obtaining the CTT, SIT optimizes it greedily to prevent the tree from collapsing to the average modal of data because there is only a single ground truth to refer to.
Particularly, we convert the ground truth to coarse ground truth by high-order velocity and the generated coarse ground truth is used to optimize the closest path to it in the CTT.
Finally, a teacher-forcing refining strategy is used to refine the top-$1$ coarse future trajectory scored and selected from the optimized CTT in training time, while the top-$K$ coarse future trajectories are selected to obtain the final multi-modal future trajectories in inference time.

We conduct extensive experiments on two popular benchmark pedestrian trajectory prediction datasets, \emph{i.e.}, ETH-UCY~\cite{eth,ucy} and Stanford Drone~\cite{sdd}. 
Despite the hand-crafted tree, the experimental results demonstrate that: 1) the proposed SIT is capable of matching or exceeding the performance of state-of-the-art methods; 2) SIT contributes to breaking the stereotype of hand-crafted methods in pedestrian trajectory prediction. Without any training, the raw ternary tree can outperform many deep neural-based methods; 3) SIT shows effective interpretability to explain pedestrians' future moving behaviors; 4) SIT shows the sufficient flexibility in long-term prediction and different best-of-$K$ predictions.

\section{Related Work}

\noindent\textbf{Pedestrian Trajectory Prediction}.
Traditionally, pedestrian trajectory prediction has been studied by hand-crafted methods. 
Since the trajectory space of pedestrians is a 2D plane, many works~\cite{discretechoicemodels,robin2009specification,ondvrej2010synthetic} split this space into multiple subspaces and then calculate the probability of each subspace based on the strong prior information.
Relying on prior knowledge, the hand-crafted methods show interpretability to explain the predicted trajectory.
Unfortunately, they are restricted in specific scenarios and difficultly generalize to more complex scenarios due to over-depending on prior knowledge.

Recently, deep learning has been applied to visual recognition~\cite{Hu:TIP2021}, action recognition~\cite{Islam:AAAI2021}, image denoising~\cite{Yu:TOG2021}, style transfer~\cite{Xu:ICCV2021}, shadow removal~\cite{Wei:CGF2019, Zhang:AAAI2020, Chen:ICCV2021}, anomaly detection~\cite{Liu:ICCV2021}, human motion prediction~\cite{Dang:ICCV2021}, as well as image and video forgery detection research~\cite{Islam:CVPR2020}. Thanks to deep learning, pedestrian trajectory prediction achieves significant progress.
As a temporal sequential learning task, many works~\cite{slstm,sgan,grouplstm,srlstm} employ the recurrent neural networks~(RNNs) or its variants~(LSTM and GRU) to capture temporal dependencies and spatial interaction. 
Considering the interaction as a spatial graph~\cite{sun2020recursive, sbigat, baidu, stgcnn, sgcn,aaai1}, the graph convolutional network~(GCN)~\cite{gcn} with a physical adjacency matrix and the attention mechanism~\cite{transformer} with a learnable adjacency matrix are used to integrate spatial interactive messages.
Moreover, some works~\cite{trajectron,Sophie,PITF,Introvert} leverage the visual information to improve the prediction performance.

Due to the multimodality of future trajectory, most works focus on the generative model to predict multi-modal future trajectories.
The CVAE-based methods~\cite{Desire,trajectron,pec} and the GAN-based methods~\cite{sgan,Sophie} map each possible future trajectory into a latent space in training time, and sample repeatedly from the latent space to obtain the multi-modal results in inference time.
In contrast, our method builds a trajectory tree on more general rules~(\emph{i.e.}, go straight, turn left and right)
to represent multi-modal future trajectories and then optimizes it to obtain the fine-grained predicted trajectory.
Thus, it is not only suitable for various scenarios, but also can provide better interpretability, stable predicted results, and sufficient flexibility in different prediction settings, verified by the experimental results. 

\noindent\textbf{Tree in Trajectory}.
Tree-based algorithms in trajectory related tasks mainly focus on path planning~\cite{trajectoryplanning}, which aims to search an acceptable path to the given destination.
LaValle {\em et al.}~\cite{rrt} propose a typical sampling-based planning approach, which extends non-holonomic constraints and supports dynamic environments as well.
Followed by that, many variants~\cite{rrt2,rrt3,rrt4} are proposed to improve the performance of path planning.
There are also tree-based trajectory prediction methods that serve motion and path planning.
Aoude {\em et al.}~\cite{aoude2011mobile} combine the closed-loop rapidly-exploring random tree~(CL-RRT)~\cite{kuwata2009real} with a Gaussian mixture model for collision avoidance and conflict detection.
Jurgenson {\em et al.}~\cite{jurgenson2019sub} divide a path into multiple sub-goals and use a divide-and-conquer process to generate a complete trajectory.
In contrast, pedestrian trajectory prediction is more challenging due to the absence of any future trajectory information.

\begin{figure}[t]
\centering
\includegraphics[width=0.8\columnwidth]{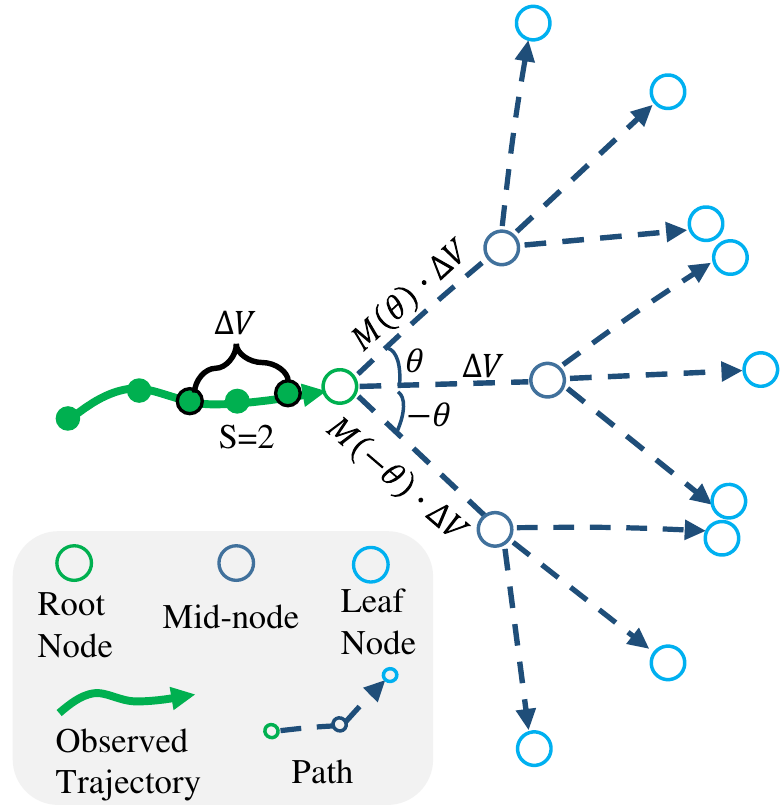} 
\caption{An example of the generating of coarse trajectory tree with the depth $d=2$ and interval $S=2$.  
The high-order velocity $\Delta V$, \emph{i.e.}, the displacement in a temporal interval $S$, is regarded as the forward split direction. The left split direction is gained by multiplication between the rotation matrix $M$ with angle $\theta$ and $\Delta V$, while the right split direction is obtained by multiplication between the rotation matrix $M$ with angle $-\theta$ and $\Delta V$. The coarse trajectory tree is generated recursively 
at each split, where each path from the root to leaf represents a coarse possible future trajectory.}
\label{fig2}
\vspace{-0.4cm}
\end{figure}

\begin{figure*}[t]
\centering
\includegraphics[width=0.98\textwidth]{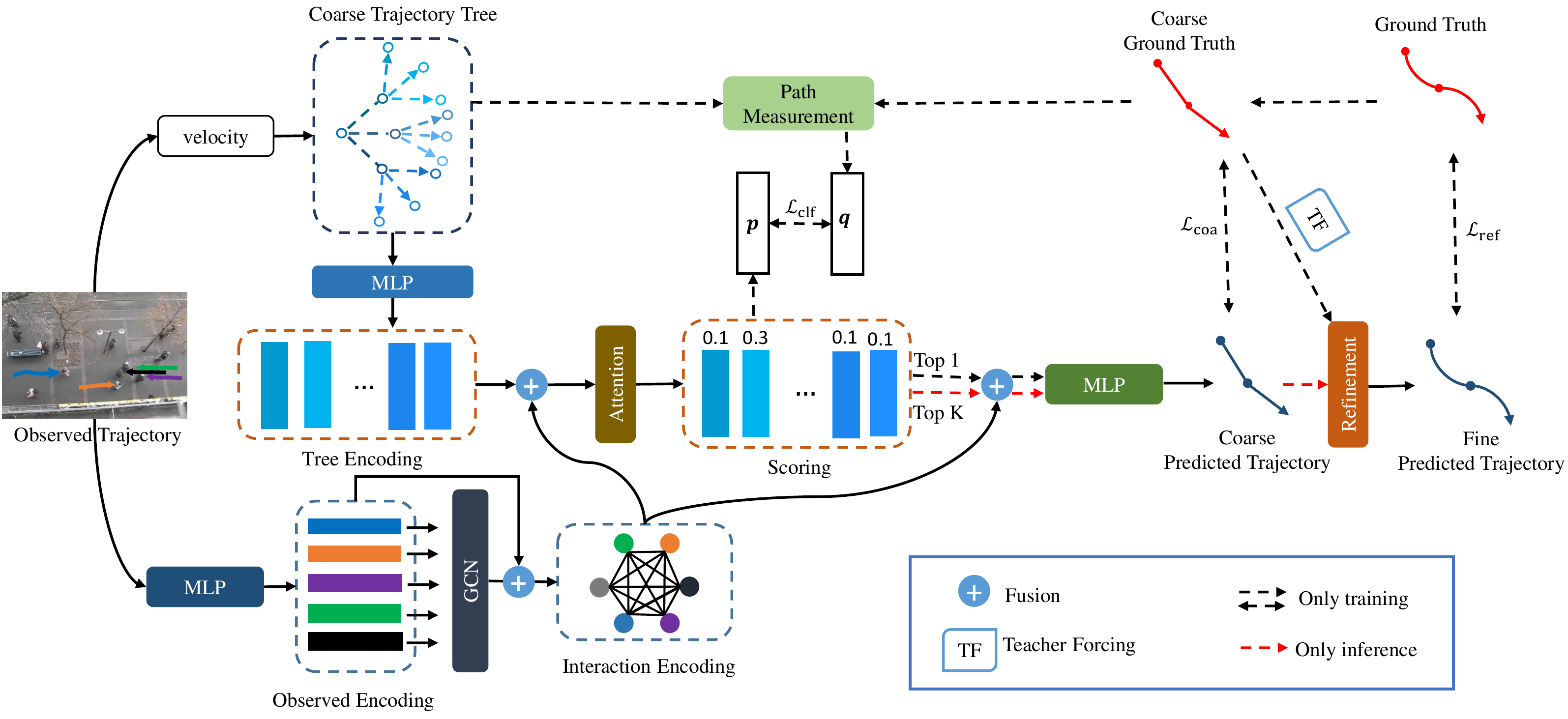} 
\caption{The overall framework of our method. The coarse trajectory tree is built firstly and then encoded by an MLP to gain the tree encoding.
Another MLP and a GCN are used to obtain the observed encoding and interaction encoding one after another from the observed trajectory.
Next, an attention mechanism is used to score each path between tree encoding and interaction encoding, and the results~(confidence vector $\text{p}$) is optimized by the index label~($\text{q}$) of the distance between each path and coarse ground truth. 
Particularly, the coarse ground truth is generated by the high-order velocity of ground truth.
Subsequently, the coarse predicted trajectory is obtained by the top-$1$ path and refined to predict fine future trajectory. 
Notably, the teacher forcing is used in refinement, which means the coarse ground truth is fed into the refinement in training time, while the coarse predicted trajectories are fed into the refinement to obtain multi-modal future trajectories in inference time.
}
\label{fig3}
\vspace{-0.5cm}
\end{figure*}

\section{Our Method}

\subsection{Problem Formulation}

Given a traffic scenario, $x_i^t$ represents the spatial coordinate of pedestrian $i$ at the time step $t$.
To collect $N$ pedestrians' coordinates from time step $1$ to $T_\text{obs}$, we can obtain the observed trajectories denoted as  $\textbf{\textit{X}}=\{\mathbf{x}_i\}_{i=1}^N$, where $\mathbf{x}_i = \{x_i^t, \}^{T_\text{obs}}_{t=1}$.
Due to the multimodality of future trajectory, there are $K$ socially-acceptable future trajectories denoted by $\mathbf{Y}=\{\textbf{\textit{Y}}_j\}_{j=1}^{K}$.
The single ground truth $\hat{\textbf{\textit{Y}}}=\{\mathbf{y}_i\}_{i=1}^N$, where $ \mathbf{y}_i = \{x_i^t\}_{t=T_\text{obs+1}}^{T_\text{pred}}, \hat{\textbf{\textit{Y}}} \in \mathbf{Y}$. 
In a real traffic scenario, the trajectory is not only affected by the pedestrians' intention but also the interaction between pedestrians at each time step $t$ denoted by $\textbf{\textit{S}}=\{s_t\}_{t=1}^{T_{obs}}$.
Briefly speaking, our objective has two parts. First, the model predicts all socially-acceptable future trajectories $\mathbf{Y}$ based on the observed trajectory $\textbf{\textit{X}}$ and interaction $\textbf{\textit{S}}$, and then selects the trajectories with high confidence to obtain the final multi-modal future trajectories.

As discussed above, the process of our method can be formulated mathematically as

\begin{equation}
\begin{aligned}
p(\hat{\textbf{\textit{Y}}} | \textbf{\textit{X}}, \textbf{\textit{S}}) = \sum_{\textbf{\textit{Y}} \in \mathbf{Y}}p(\textbf{\textit{Y}}|\textbf{\textit{X}}, \textbf{\textit{S}}) p(\hat{\textbf{\textit{Y}}}|\textbf{\textit{Y}}, \textbf{\textit{X}}, \textbf{\textit{S}}),
\end{aligned}
\label{eq1}
\end{equation}
where $p(.|.)$ is a discrete conditional distribution because $\mathbf{Y}$ is represented by a tree.

Under this formulation, previous works embed $\mathbf{Y}$ into an implicit continuous space~\cite{sgan,pec} by specific generative models,
where the selecting process is directly replaced with the sampling repeatedly from the learned latent space in reference time.
In contrast, our method embeds $\mathbf{Y}$ into a discrete structured space, which is specifically represented by a ternary tree. 
Since the ternary tree is not influenced by the average modal of data, each trajectory~(path) contained in the tree can keep its individual moving behavior, and thus provide well interpretability and meanwhile not fall into the frequent modal. 
Moreover, the operation of selection could generate stable predicted results compared with sampling repeatedly. 

The overall framework of our method is illustrated in Figure~\ref{fig3}.
Specifically, the coarse trajectory tree is built firstly to generate the coarse discrete structured space $\textbf{Y}_\text{coarse}$ and then encoded by an MLP to gain the tree encoding.
Meanwhile, the observed trajectory and spatial interaction are encoded to obtain the observed encoding and interaction encoding one after another.
Next, the tree and interaction are fused to score each path in the coarse trajectory tree by an attention mechanism, and then the obtained confidence vector $\mathbf{p}$ is optimized supervised by the label $\mathbf{q}$, which is obtained by the path measurement between each path and the coarse ground truth. 
Particularly, the coarse ground truth is generated by the high-order velocity of ground truth.
Subsequently, the $\textbf{Y}_\text{coarse}$ is optimized greedily by the
path with the highest confidence to generate the coarse predicted trajectory supervised by the coarse ground truth.
Finally, a refining operation is employed on the coarse predicted trajectory to gain the fine-grained trajectory with the teacher forcing, which means the refining operation uses coarse ground truth for refinement in training time, while the coarse predicted trajectories with top-k confidences are used to refine for multi-modal future trajectory prediction in reference time.

\subsection{Trajectory Prediction with Tree}
\noindent\textbf{Coarse Trajectory Tree}.
The fundamental operation for our method is to build the coarse trajectory tree, which refers to the generating of a ternary tree as preceding discussion.
The whole process is considered as a recursive split in three directions~(ternary tree) at each time step.
Due to the temporal dependency of trajectory, the velocity vector of the observed trajectory is used to get the direction of forwarding split~(go straight). 
In particular, the directions of left split~(turn left) and right split~(turn right) are gained by positive and negative rotation of the velocity vector with a specific angle, respectively.
As shown in Figure~\ref{fig2}, 
given the location of observed trajectories $\mathbf{x}_i$~(green arrowed line) for the pedestrian $i$, 
we can obtain the corresponding velocities denoted 
$\textbf{v}_i = \{ v_i^t\}_{t=1}^{T_\text{obs}}$, by the displacement from one time step to next time step.
Note that we assume the pedestrian keeps still at the first time step, namely the $\{v_i^1\}_{i=1}^N=\mathbf{0}$.
Since the complexity of tree grows exponentially as the depth~($d$) increases, \emph{e.g.}, assuming the predicted length  $T = T_\text{pred} - T_\text{obs} = 12$, we will generate a ternary tree with the $d=12$ and it has $3^{12}$ paths if the split is taken at each time step.
To balance the complexity and the coverage of the tree, the tree splits multi-time steps once instead of split time step by time step.
Therefore, we set a specific temporal interval~($1 \leq S \leq T$) and the high-order velocity~($\Delta V$), gained by summing all velocity vectors in the last $S$ of observed trajectory, is considered as the split direction of forwarding direction.
The rotated angle~($\theta$) is used to generate the directions of the left and right split.
Finally, we will generate a ternary tree, \emph{i.e.}, coarse trajectory tree, with the depth $d=\lceil T / S \rceil$ after a recursive process.
Upon obtaining the coarse trajectory tree, each path from the root to leaf represents a coarse possible future trajectory, and thus the coarse discrete structured space $\mathbf{Y}_\text{coarse}$ could be composed of all paths in the coarse trajectory tree. 

\noindent\textbf{Trajectory Encoding}.
The future trajectory is not only affected by the internal motion information but the interactive states with other pedestrians.
In this paper, we mainly focus on evaluating the effectiveness of the tree for pedestrian trajectory prediction. 
We use an simply multilayer perceptron~(MLP) to encode the observed trajectory $\textbf{\textit{X}}$ into observed encoding denoted $\mathbf{F}_\text{x}$.
Another MLP is applied to encode the path of $ \mathbf{Y}_\text{coarse}$ into tree encoding 
denoted $\mathbf{F}_\text{tree} = \{ \mathbf{f}_i \}_{i=1}^M$, where the $M$ is the size of  $ \mathbf{Y}_\text{coarse}$.
In addition, a graph convolutional network~(GCN)~\cite{gcn} implemented by the self-attention~\cite{transformer} without the positional encoding is used to model the interaction encoding denoted $\mathbf{F}_\text{s}$.

\noindent\textbf{Scoring and Selection}.
After obtaining the coarse discrete structured space $\mathbf{Y}_\text{coarse}$, we need to optimize it to obtain more precise trajectory.
To keep the interpretability of tree, we use a two-stage training strategy, in which the path in the $\mathbf{Y}_\text{coarse}$ is first scored and then those with high confidence are selected to optimize.

To score the path, 
we model the attention scores between the interaction encoding $\mathbf{F}_\text{s}$ and the tree encoding $\mathbf{F}_\text{tree}$ as the confidence vector $\mathbf{p}$ , \emph{i.e.},
\begin{equation}
\begin{aligned}
\mathbf{p} &= \operatorname{Softmax}(\phi(\mathbf{\mathbf{F}_\text{s})}\psi(\mathbf{F}_\text{tree})^{\text{T}}), \\
\end{aligned}
\label{eq4}
\end{equation}
where $\phi$ and $\psi$ are the linear projections, $\text{T}$ is the transpose.

After that, since the closest path with ground truth can provide rough explanation about the moving behavior of ground truth, we expect it gains the highest confidence. 
Thus, a path measurement is employed to measure the distance between each path and the ground truth, and the location index in the coarse trajectory tree of the closest one is considered as the label $\mathbf{q}$ to supervise the scoring operation.
Particularly, since the real trajectory of pedestrian is zigzag,  we convert the ground truth $\hat{\textbf{\textit{Y}}}$ to its coarse version $\hat{\textbf{\textit{Y}}}_\text{coarse}$ to simplify the optimization. Namely, $\mathbf{q}$ is generated by measuring the distance between each path and $\hat{\textbf{\textit{Y}}}_\text{coarse}$.
Similar with coarse trajectory tree, $\hat{\textbf{\textit{Y}}}_\text{coarse}$ is generated by dividing the ground truth into multiple equilong segments with temporal interval $S$ and then connecting the break point as illustrated ground truth and coarse ground truth in Figure~\ref{fig3}.
The distance of path measurement is the mean of L-2 distance between each break point and corresponding point in the path.
The loss function can be given by

\begin{equation}
\begin{aligned} 
\mathcal{L}_\text{clf} = \mathcal{L}_
\text{CE}(\mathbf{p}, \mathbf{q}),
\end{aligned}
\label{eq5}
\end{equation}
where the $\mathcal{L}_\text{CE}$ is the cross entropy loss.

\noindent\textbf{Greedy Optimization}. 
Due to the single provided ground truth, the model will collapse into frequent modal of data if we force multiple paths to approach the ground truth. 
To obtain multi-modal future trajectory, we optimize the path greedily, which means the path with highest confidence is used to optimize the $\mathbf{Y}_\text{coarse}$.
Specifically, the tree encoding $\mathbf{f}_\text{*}$ of the path with highest confidence fused with the interaction encoding $\mathbf{F}_\text{s}$ is fed into an MLP to obtain the coarse predicted trajectory $\textbf{\textit{Y}}^{'}$.
The objective function of the greedy optimization is shown by 
\begin{equation}
\begin{aligned}
\mathcal{L}_\text{coarse} = \mathcal{L}_
\text{reg}(\textbf{\textit{Y}}^{'}, \hat{\textbf{\textit{Y}}}_\text{coarse}),
\end{aligned}
\label{eq3}
\end{equation}
where $\mathcal{L}_\text{reg}$ is the Huber loss.

\noindent\textbf{Trajectory Refining}.
The final step of our method refine the coarse predicted trajectory $\textbf{\textit{Y}}^{'}$ to obtain fine-grained trajectory $\textbf{\textit{Y}}_\text{fine}$.
To ensure the optimization of refining correctly especially in the early stage of training, we use a teacher-forcing~\cite{teacherforcing} strategy in training time. 
Namely, the closest coarse trajectory $\textbf{\textit{Y}}^{'}$ is replaced with the coarse ground truth $\textbf{\textit{Y}}_\text{coarse}$ to regress the final fine-grained trajectory.
The loss function of trajectory refining is represented by

\begin{equation}
\begin{aligned}
\mathcal{L}_\text{ref} = \mathcal{L}_
\text{reg}(\textbf{\textit{Y}}_\text{coarse}, \textbf{\textit{Y}}),
\end{aligned}
\label{eq6}
\end{equation}
where $\mathcal{L}_\text{reg}$ is the Huber loss, $\textbf{\textit{Y}}$ is the ground truth.

\noindent\textbf{Training and Inference}. We train the proposed method in an end-to-end way. The total loss is

\begin{equation}
\begin{aligned}
\mathcal{L} = \lambda_1\mathcal{L}_\text{coarse} + 
\lambda_2\mathcal{L}_\text{clf} + 
\lambda_3\mathcal{L}_\text{ref},
\end{aligned}
\label{eq7}
\end{equation}
where $\lambda_1$, $\lambda_2$ and $\lambda_3$ are used to balance the training process.

At the inference step, we select the top-$K$ predicted trajectories according to the attention scores, and those are refined to obtain the final $K$ fine-grained trajectories, \emph{i.e.}, multi-modal future trajectories.

\noindent\textbf{Implementation Details}.
To implement the proposed method, all encoding modules are implemented by a 3-layer MLP with the PRelu non-linearity.
We split the coarse trajectory tree three times and generates $27$ paths. Other hyper-parameters of this tree are recorded in Appendix due to space limitation.
All the coefficients $\lambda_1$, $\lambda_2$, and $\lambda_3$ of total loss are set to 1.

\section{Experimental Analysis}

\noindent\textbf{Datasets}. 
To evaluate the effectiveness of our method, we conduct extensive experiments on two widely used datasets, \emph{i.e.}, ETH-UCY~\cite{eth,ucy} and Stanford Drone Dataset~(SDD)~\cite{sdd}, in pedestrian trajectory prediction.
ETH-UCY includes five scenes: ETH, HOTEL UNIV, ZARA1 and ZARA2, and the coordinate of trajectory is recorded in world coordinate system with the meter as the unit. SDD contains $20$ scenes and the coordinate of trajectory is recorded in pixel coordinate system with the pixel as the unit.
For ETH-UCY, we follow the leave-one-out strategy~\cite{sgcn} for training and evaluation, which the model is trained on four scenes and evaluated on the rest of the scene.
For SDD, we use prior train-test split~\cite{pec} for evaluation.

Following the common setting~\cite{sgcn}, we segment the trajectory sequences into $8s$ trajectory segments by sliding, in which the observed trajectory is $3.2s$ and the future trajectory is the rest $4.8s$, with a time step of $0.4s$.

\noindent\textbf{Metrics}. Following the common practice~\cite{sgan}, we adopt two widely used metrics to evaluate the performance of the predicted trajectory. Average Displacement Error~(ADE) computes the average L-2 distance between predicted trajectory location and the ground truth location.
Final Displacement Error~(FDE) calculates the L-2 distance between the predicted trajectory at the last time step location and the corresponding group truth location. 
To measure the ADE and FDE of our method, we follow the previously commonly used measurement  that selects $K$ predicted trajectories and reports the performance of closest trajectory.

\begin{table*}[t]
\centering
\begin{tabular}{ccc|ccccc|c}
\toprule
Model   &   Venue & Year          & ETH & HOTEL & UNIV & ZARA1 & ZARA2 & AVG      \\
\midrule
Vanilla LSTM & CVPR & 2016   & 1.09/2.41 & 0.86/1.91 & 0.61/1.31 & 0.41/0.88 & 0.52/1.11 & 0.70/1.52 \\
Social LSTM & CVPR & 2016   & 1.09/2.35 & 0.79/1.76 & 0.67/1.40 & 0.47/1.00 & 0.56/1.17 & 0.72/1.54 \\
Desire & CVPR & 2017   & 0.73/1.65 & 0.30/0.59 & 0.60/1.27 & 0.38/0.81 & 0.31/0.68 & 0.46/1.00 \\
Sophie & CVPR & 2019    & 0.70/1.43 & 0.76/1.67 & 0.54/1.24 & 0.30/0.63 & 0.38/0.78 & 0.51/1.15 \\
GAT & NeurIPS & 2019  & 0.68/1.29 & 0.68/1.40 & 0.57/1.29 & 0.29/0.60 & 0.37/0.75 & 0.52/1.07 \\
Social-BIGAT & NeurIPS & 2019 & 0.69/1.29 & 0.49/1.01 & 0.55/1.32 & 0.30/0.62 & 0.36/0.75 & 0.48/1.00 \\
STAR & ECCV & 2020 & \textbf{0.36}/0.65 & 0.17/0.36 & 0.31/0.62 & 0.26/0.55 & 0.22/0.46 & 0.26/0.53 \\
PECNet & ECCV & 2020 & 0.47/0.87 & 0.18/0.24 & 0.35/0.60 & 0.22/0.39 & 0.17/0.30 & 0.29/0.48 \\
SGCN      & CVPR & 2021       & 0.63/1.03 & 0.32/0.55 & 0.37/0.70 & 0.29/0.53 & 0.25/0.45 & 0.37/0.65 \\
DMRGCN      & AAAI & 2021       & 0.60/1.09 & 0.21/0.30 & 0.35/0.63 & 0.29/0.47 & 0.25/0.41 & 0.34/0.58 \\
\hline
SGAN  & CVPR & 2018   & 0.87/1.62 & 0.67/1.37 & 0.76/1.52 & 0.35/0.68 & 0.42/0.84 & 0.61/1.21 \\ Social-STGCNN & CVPR & 2020 & 0.64/1.11 & 0.49/0.85 & 0.44/0.79 & 0.34/0.53 & 0.30/0.48 & 0.44/0.75 \\
TPNMS      & AAAI & 2021       & 0.52/0.89 & 0.22/0.39 & 0.55/1.13 & 0.35/0.70 & 0.27/0.56 & 0.38/0.73 \\
\hline
SIT~(Ours) & AAAI & 2022 & 0.39/\textbf{0.62} & \textbf{0.14/0.22} & \textbf{0.27/0.47} & \textbf{0.19/0.33} & \textbf{0.16/0.29} & \textbf{0.23/0.38} \\
Raw Tree~($d$=0)      & AAAI & 2022   & 0.99/2.23  & 0.32/0.61 & 0.52/1.16 & 0.43/0.96 & 0.32/0.72 & 0.51/1.13 \\
Raw Tree~($d$=1)  & AAAI & 2022 & 0.91/2.00 & 0.27/0.51 & 0.43/0.94 & 0.35/0.75 & 0.26/0.56 & 0.44/0.95 \\
Raw Tree~($d$=2)  & AAAI & 2022 & 0.86/1.85 & 0.25/0.46 & 0.41/0.90 & 0.31/0.65 & 0.23/0.51 & 0.41/0.87 \\
Raw Tree~($d$=3)  & AAAI & 2022 & 0.82/1.64 & 0.24/0.40 & 0.38/0.77 & 0.29/0.53 & 0.22/0.43 & 0.39/0.75 \\
\bottomrule
\end{tabular}
\vspace{-0.1cm}
\caption{Comparison with baselines on the ETH-UCY using ADE/FDE, which are measured in meters. The lower the better. The models SGAN, Social-STGCNN and TPNMS are the closest methods compared with our raw tree.}
\label{table1}
\vspace{-0.5cm}
\end{table*}

\begin{table}[t]
    \centering
    \begin{tabular}{cc|cc}
    \toprule
    Models & K  & ADE & FDE\\
    \midrule
        SoPhie & 20 & 16.27 & 29.38 \\
        SGAN   & 20  & 27.24 & 41.44 \\
        Desire & 5  & 19.25 & 34.05 \\
        CF-VAE & 20 & 12.60 & 22.30 \\
        P2TIRL & 20 & 12.58 & 22.07 \\
        SimAug & 20 & 10.27 & 19.71 \\
        PECNet & 5  & 12.79 & 25.98 \\
        PECNet & 20 & 9.96  & 15.88 \\
        \hline
        SIT~(Ours) & 25 & 8.93 & 14.97 \\
        SIT~(Ours) & 20 &\textbf{9.13} &  \textbf{15.42} \\
        SIT~(Ours) & 15 &9.48 &  16.48 \\
        SIT~(Ours) & 10 &10.34 &  18.50 \\
        SIT~(Ours) & 5 &12.30 &  23.17 \\
        \bottomrule
    \end{tabular}
    \caption{Comparison with baselines on Stanford Drone using ADE and FDE. The metrics are measured in pixels.}
    \label{table2}
    \vspace{-0.5cm}
\end{table}

\begin{table}[t]
\normalsize
\centering
\begin{tabular}{c|c|cc}
\toprule
$T_\text{pred}$ & Models  & ADE & FDE      \\
\midrule
\multirow{4}{*}{16} & SGAN  & 2.16 & 3.96 \\
&Social-STGCNN  & 0.54 & 1.05 \\
&PECNet & 2.89 & 2.63 \\
& SIT  & \textbf{0.49}&\textbf{1.01} \\
\hline
\multirow{4}{*}{20} & SGAN  & 2.40 & 4.52 \\
& Social-STGCNN  & 0.71&1.30 \\
& PECNet  & 3.02&2.55 \\
& SIT  & \textbf{0.55}&\textbf{1.12} \\
\hline
\multirow{4}{*}{24}& SGAN  & 2.79 & 4.66 \\
& Social-STGCNN  & 0.92&1.76 \\
& PECNet   & 3.16&2.53 \\
& SIT   & \textbf{0.68}&\textbf{1.22} \\
\bottomrule
\end{tabular}
\caption{Long-term prediction on ETH-UCY.}
\label{tab:long_term_pred}
\vspace{-0.6cm}
\end{table}

\begin{table}[t]
\normalsize
\centering
\begin{tabular}{c|c|cc}
\toprule
$K$ & Models  & ADE & FDE       \\
\midrule
\multirow{2}{*}{5}
& PECNet  & 0.64 & 1.21 \\
& SIT  & \textbf{0.35}&\textbf{0.65} \\
\hline
\multirow{2}{*}{10}
& PECNet  & 0.64&1.14 \\
& SIT  & \textbf{0.27}&\textbf{0.49} \\
\hline
\multirow{2}{*}{15}
& PECNet   & 0.64 & 1.11 \\
& SIT   & \textbf{0.24}&\textbf{0.42} \\
\bottomrule
\end{tabular}
\caption{Different best-of-$K$ predictions on ETY-UCY.}
\label{tab:best-of-k}
\vspace{-0.4cm}
\end{table}

\subsection{Quantitative Analysis}
We conduct extensive experiments to evaluate the effectiveness of SIT in prediction accuracy,  raw tree prediction, long-term prediction, and different best-of-$K$ predictions.
More experiments are reported in Appendix due to space limitation. 

\noindent\textbf{Performance on ETH-UCY}.
The results are given in Table~\ref{table1}, 
which are evaluated by ADE and FDE.
Although our proposed SIT is based on a hand-crafted tree, the results indicate that our SIT outperforms all the competing methods on both ADE and FDE on average.
Specifically,
for ADE, our SIT surpasses the previous best method STAR~\cite{star} by 11.5\% on average.
For FDE, our SIT outperforms the previous best method PECNet~\cite{pec} by a margin of 20.8\% on average.
The performances on both ADE and FDE underline the effectiveness of the tree in pedestrian trajectory prediction.

\noindent\textbf{Prediction with Raw Tree}. 
We conduct a specific experiment to testify the tree is suitable for pedestrian trajectory prediction even without any training.
The raw tree~(\emph{i.e.}, coarse trajectory tree) is built only based on the prior information, \emph{i.e.}, velocity, and it is directly used to compare with other deep learning-based methods as shown in Table~\ref{table1}.
The raw tree is tested with different depth $d$, which is set to $0, 1, 2,$ and $3$, respectively.
Note that $d=0$ means the pedestrians keep going straight along the direction of the last time step of the observed trajectory.
Since the tree is ternary, we can obtain $3^d$ trajectories for each $d$.
The experimental results demonstrate our raw tree can match the deep learning-based methods, \emph{i.e.}, Social-STGCNN~\cite{stgcnn} and TPNMS~\cite{aaai2}.
Interestingly, the raw tree with $d=0$, \emph{i.e.}, only a trajectory keeps going straight, exceeds SGAN~\cite{sgan} which uses best-of-$20$ to report metrics.
This phenomenon indicate our raw tree could cover effective space of future trajectory even it is built by hand.
Based on a general rule,~(\emph{i.e.}, go straight, tree left and turn right ), the raw tree can generate effective trajectory that is more suitable for various scenarios and thus obtains better performance.

\noindent\textbf{Performance on SDD}. 
As shown in Table~\ref{table2},
our method outperforms previous state-of-the-art methods~\cite{pec} on both ADE and FDE.
It shows higher feasibility in pedestrian trajectory prediction. 
The trajectory prediction with various $k$ shown in Table~\ref{table2} also indicates the effectiveness of our SIT.
More experimental results please see Appendix.

\noindent\textbf{Long-term Prediction}. We conduct experiments on long-term prediction, which input the observed trajectory with normal~($3.2s$ and $8$ time steps) length, the longer future trajectory will be predicted.
\begin{figure*}[t]
\centering
\includegraphics[width=\textwidth]{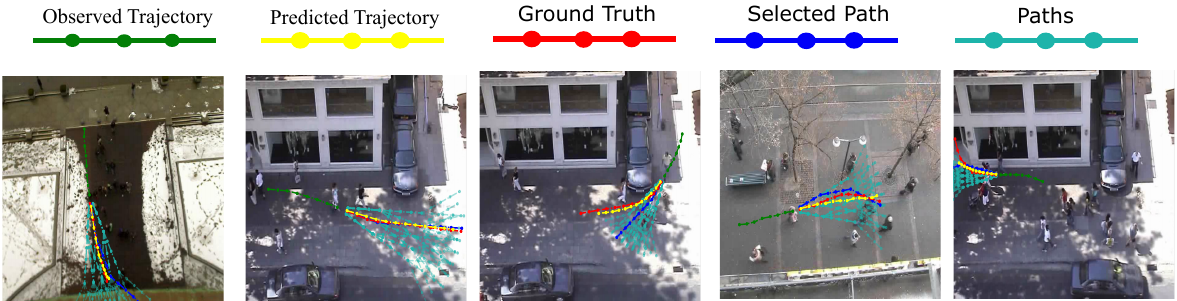} 
\caption{Visualization of selected path and refined trajectory.}
\label{fig5}
\vspace{-0.4cm}
\end{figure*}
In this experiment, we set the longer future trajectory to $6.4s$~($16$ time steps), $8.0s$~($20$ times steps), and $9.6s$~($24$ time steps), respectively.
To indicate the flexibility of our SIT on long-term prediction by comparing against other baselines, we reproduce the LSTM and GAN-based method SGAN~\cite{sgan}, the graph-based method Social-STGCNN~\cite{stgcnn} and the CVAE-based method PECNet~\cite{pec} by their official released codes
\footnote[3]{
SGAN:\url{https://github.com/agrimgupta92/sgan}, 

PECNet:\url{https://github.com/HarshayuGirase/PECNet},

Social-STGCNN:\url{https://github.com/abduallahmohamed/Social-STGCNN}
}
to predict long-term future trajectory, respectively.
Note that the PECNet is reproduced by the data loader of Social-STGCNN because the data loader of PECNet can not change the predicted length.
As shown in Table~\ref{tab:long_term_pred}, our SIT outperforms all competing methods on all long-term predicted lengths. 
Notably, the improvements are gradually increasing as the predicted length elongates.
The underlying reason could be that the built tree provides effective ``candidates"~(paths) that are convenient for the optimization of deep neural network.

\noindent\textbf{Different best-of-$K$ predictions}. 
Due to the multi-modal of future trajectory, related works use the best-of-$K$ to report the quantified metrics. 
Namely, $K$~(usually $K=20$) future trajectories are predicted, while only the closest trajectory is used to report.
To further testify the flexibility of our SIT, we conduct experiments on different best-of-$K$ predictions, where we set $K= 15, 10$, and $5$, respectively.
We also reproduce the state-of-the-art method PECNet for comparison.
Since PECNet does not provide pretrained mode, we compare with it by above reproduced model. The experimental results are presented in Table~\ref{tab:best-of-k}. It indicates that our SIT achieves significant performance on all experimental settings.
Interestingly, our SIT with a small $K$, \emph{e.g.}, SIT-5, still outperforms Social-STGCNN~\cite{stgcnn} with $K=20$, as compared between Table~\ref{table1} and Table~\ref{tab:best-of-k}.

\subsection{Ablation Study}

We conduct ablative experiments to isolate the performance contribution of each component of our method.
The relevant components include the teacher forcing~(TF), classification task~(CLF), coarse ground truth~(CGT), and the interaction encoding~(IE). 
Specifically, to represent their effectiveness, the TF is verified by replacing the coarse ground truth with the path with the highest confidence~(1), 
the CLF is verified by removing the loss function $\mathcal{L}_\text{clf}$~(2),
the CGT is verified by removing the TF and CLF together~(3),
and the IE is verified by removing the attention encoding~(4).
Table~\ref{table6} presents the full method~(5) achieves the best performance, which clearly validates the effectiveness of each component.

\subsection{Qualitative Analysis}

\noindent\textbf{Interpretability}. 
The interpretability of our SIT mainly refers to the path in the tree that can provide a good explanation of future moving behaviors, \emph{e.g.}, go straight and then turn right. 
To show our SIT is capable of selecting the closest path with ground truth, we make statistics in the testing set to record the rate that selecting the closest path in different best-of-$K$ predictions.
The closest path is selected by the minimum FDE with the coarse ground truth. Relevant results of minimum ADE are shown in Appendix.
As shown in Table~\ref{table5}, our SIT shows significant accuracy~(88.47\% to 97.38\%) in standard best-of-$20$ prediction.
For the lower accuracy of top-$1$, the reason is SIT encourages multi-modal prediction, which does not welcome the single prediction.

\begin{table}[t]
\centering
\begin{tabular}{c|ccccc}
\toprule
K           & ETH & HOTEL & UNIV & ZARA1 & ZARA2   \\
\midrule
1   & 28.72\% & 41.59\% & 9.81\% & 14.29\% & 9.18\% \\
5   & 58.56\% & 58.68\% & 36.29\% & 59.52\% & 39.17\% \\
10   & 80.11\% & 73.59\% & 59.20\% & 79.49\% & 62.50\% \\
15   & 88.39\% & 87.36\% & 76.46\% & 90.19\% & 79.92\% \\
20   & 93.92\% & 94.39\% & 88.47\% & 97.38\% & 89.73\% \\
\bottomrule
\end{tabular}
\caption{Top-$K$ accuracy of selecting the closest path of tree.}
\label{table5}
\vspace{-0.4cm}
\end{table}

\begin{table}[t]
\centering
\begin{tabular}{ccccc|cc}
\toprule
&TF & CLF   & CGT & IE & ADE & FDE   \\
\midrule
(1)&\XSolidBrush   & \checkmark   & \checkmark  & \checkmark & \textbf{9.13} & 15.57 \\
(2)&\XSolidBrush   & \XSolidBrush   & \checkmark  & \checkmark & 9.52 & 16.69 \\
(3)&\XSolidBrush   & \XSolidBrush   & \XSolidBrush  & \checkmark & 13.43  & 25.09 \\
(4)&\checkmark   & \checkmark   & \checkmark  & \XSolidBrush & 9.53  & 16.01 \\
(5)&\checkmark   & \checkmark   & \checkmark  & \checkmark & \textbf{9.13}  & \textbf{15.42} \\
\bottomrule
\end{tabular}
\caption{Ablation study on SDD.}
\label{table6}
\vspace{-0.6cm}
\end{table}

\noindent\textbf{Visualization}. 
To further illustrate the interpretability of our SIT, we visualize the selected path in real traffic scenarios.
As presented in Figure~\ref{fig5}, from the left to right, the images represent the pedestrians go straight and then turn left, keep going straight, keep turn right, go straight and then turn right and keep turn right.
Our SIT can select the path with similar behaviors of the ground truth and then refines it to gain a precisely predicted trajectory.
For more visualizations please see Appendix.

\section{Conclusion}
We propose a simple yet effective tree-based method, named Social Interpretable Tree~(SIT) to predict the multi-modal future trajectories.
Compared with previous methods that embed the multi-modal future trajectories into a continuous latent space, we embed them into a discrete structured space, \emph{i.e.}, a ternary tree. 
In our method, a coarse trajectory tree is first built and then a coarse-to-fine strategy is used to obtain the final multi-modal future trajectories. 
Experimental results on ETH-UCY and Stanford Drone Dataset validate the effectiveness of our SIT in standard prediction, long-term prediction, different best-of-$K$ predictions, and interpretability.
Furthermore, the raw tree without any training outperforms even many deep learning-based methods.

\section{Acknowledgments}

This work was supported partly by National Key R\&D Program of China under Grant 2018AAA0101400, NSFC under Grants 62088102, 61976171, and 62106192, China Postdoctoral Science Foundation under Grant 2020M683490, Natural Science Foundation of Shaanxi Province under Grant 2021JQ-054, and Fundamental Research Funds for the Central Universities under Grant XTR042021005.

\bibliography{aaai22}

\section{Appendix}

\subsection{Hyper-parameters of tree}

We split three times to build the coarse trajectory tree for each pedestrian. The split angles are marked by $\theta_1, theta_2$ and $\theta_3$ as represented in Table~\ref{tree_hype}. The detailed process of the building of coarse trajectory tree has been released in our \textit{official code: \url{https://github.com/lssiair/SIT}}.

\begin{table}[ht!]
    \centering
    \begin{tabular}{cccc}
    \toprule
    Dataset & $\theta_1$ & $\theta_2$ & $\theta_3$ \\
    \midrule
        ETH & 4/$\pi$& 6/$\pi$ & 4/$\pi$ \\
        HOTEL & 6/$\pi$&  6/$\pi$ & 4/$\pi$  \\
        UNIV & 4/$\pi$ &  6/$\pi$ & 4/$\pi$ \\
        ZARA1 & 12/$\pi$&  6/$\pi$ & 4/$\pi$  \\
        ZARA2 & 6/$\pi$&  6/$\pi$ & 4/$\pi$  \\
        SDD & 4/$\pi$&  6/$\pi$ & 4/$\pi$ \\
    \bottomrule
    \end{tabular}
    \caption{Hyper-parameters of the coarse trajectory tree.}
    \label{tree_hype}
\end{table}

\subsection{More Quantitative Analysis on Stanford Drone Dataset}
\noindent\textbf{Prediction with Raw Tree}.
We conduct specific experiments to evaluate whether the raw tree~(coarse trajectory tree) is suitable for pedestrian trajectory prediction on Stanford Drone Dataset~(SDD)~\cite{sdd}.
The raw tree is tested with different depth $d$, which is set to $0, 1, 2,$ and $3$, respectively.
Since the tree is ternary, we can obtain $3^d$ trajectories for each $d$.
The results are shown in Table~\ref{table1}, where the raw tree with depth $d=3$ outperforms the closest deep learning-based method Sophie~\cite{Sophie}.  The raw tree with depth $d=0$, namely a single trajectory represented going straight along the direction of last time of observed trajectory, exceeds the deep learning-based method SGAN~\cite{sgan}, which uses the Best-of-20 to report the ADE and FDE.
\begin{table}[ht!]
    \centering
    \begin{tabular}{ccc|cc}
    \toprule
    Model & Venue & Year & ADE & FDE\\
    \midrule
        Social-LSTM & CVPR& 2016 & 31.19 & 56.97 \\
        SGAN & CVPR& 2018 & 27.23 & 41.44 \\
        MATF & CVPR& 2019 & 22.59 & 33.53 \\
        Desire & CVPR& 2017 & 19.25 & 34.05 \\
        Sophie & CVPR& 2019 &16.27 & 29.38 \\
        SimAug & ECCV& 2020 & 10.27 & 19.71 \\
        PECNet & ECCV& 2020 & 9.96 & 15.88 \\
        \hline
        SIT~(Ours) & AAAI & 2022 &\textbf{9.13} & \textbf{15.42} \\
        Raw Tree~($d=0$) & AAAI & 2022 & 19.74 & 40.04  \\
        Raw Tree~($d=1$) & AAAI & 2022 & 16.93 & 33.60  \\
        Raw Tree~($d=2$) & AAAI & 2022 & 16.10 & 31.16 \\
        Raw Tree~($d=3$) & AAAI & 2022 & 15.27 & 27.06  \\
        \bottomrule
    \end{tabular}
    \caption{Comparison with baselines on Stanford Drone using ADE and FDE. The metrics are measured in pixels.}
    \label{table1}
\end{table}

\noindent\textbf{Different Best-of-$K$ predictions}.
we conduct experiments on different best-of-$K$ predictions on Stanford Drone~\cite{sdd}, where we set $K= 15, 10$, and $5$, respectively. The experimental results are shown in Table~\ref{diff_best_of_k}, where our SIT outperforms all compared methods on each $K$. 

\begin{table}[ht!]
    \centering
    \begin{tabular}{c|c|cc}
    \toprule
    $K$&Model & ADE&FDE\\
    \midrule
    \multirow{3}{*}{15} &    PECNet & 11.33&18.65 \\
    & SIT & \textbf{9.48} & \textbf{16.45} \\
    \hline
    \multirow{3}{*}{10} & PECNet & 12.39 & 21.37 \\
    & SIT & \textbf{10.34} & \textbf{18.50} \\
    \hline
     \multirow{3}{*}{5} & PECNet & 14.62 & 27.36 \\
        & SIT & \textbf{12.30} & \textbf{23.17} \\
        \bottomrule
    \end{tabular}
    \caption{Different best-of-K predictions on Standford Drone using ADE and FDE.}
    \label{diff_best_of_k}
\end{table}

\section{Ablation Study}
\begin{figure}[ht!]
\centering
\includegraphics[width=\columnwidth]{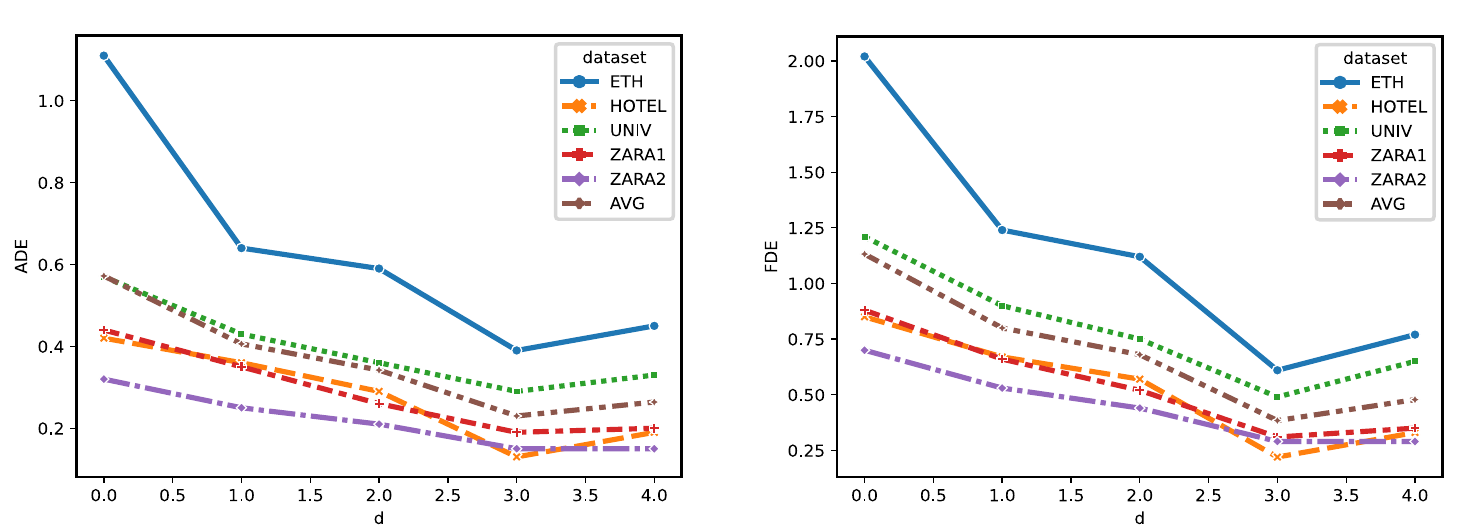} %
\caption{Illustration of the performance variation with different depths on ADE and FDE.}
\label{depth_vis}
\end{figure}

\noindent\textbf{Performance on Different Depths of Tree}.
To analyze how the performance changes with different depths of our coarse trajectory tree, we conduct ablative experiments, which set the depth $d$ to $0, 1, 2, 3,$ and $4$, respectively. For each depth $d$, we can obtain $3^d$ paths.
As shown in Figure~\ref{depth_vis},
the coarse trajectory tree occurs a depth-performance dilemma as the tree deepens constantly, which means the performance gradually changes better from 0 to 3, while it gradually changes worse from 3 to 4.
The model achieves the best performance when $d=3$.
The possible reason is that the model with the deeper tree learning to select the closest path with ground truth is difficult than the model with the shallow tree because the number of the selected path is fixed.

\begin{figure}[ht!]
\centering
\includegraphics[width=\columnwidth]{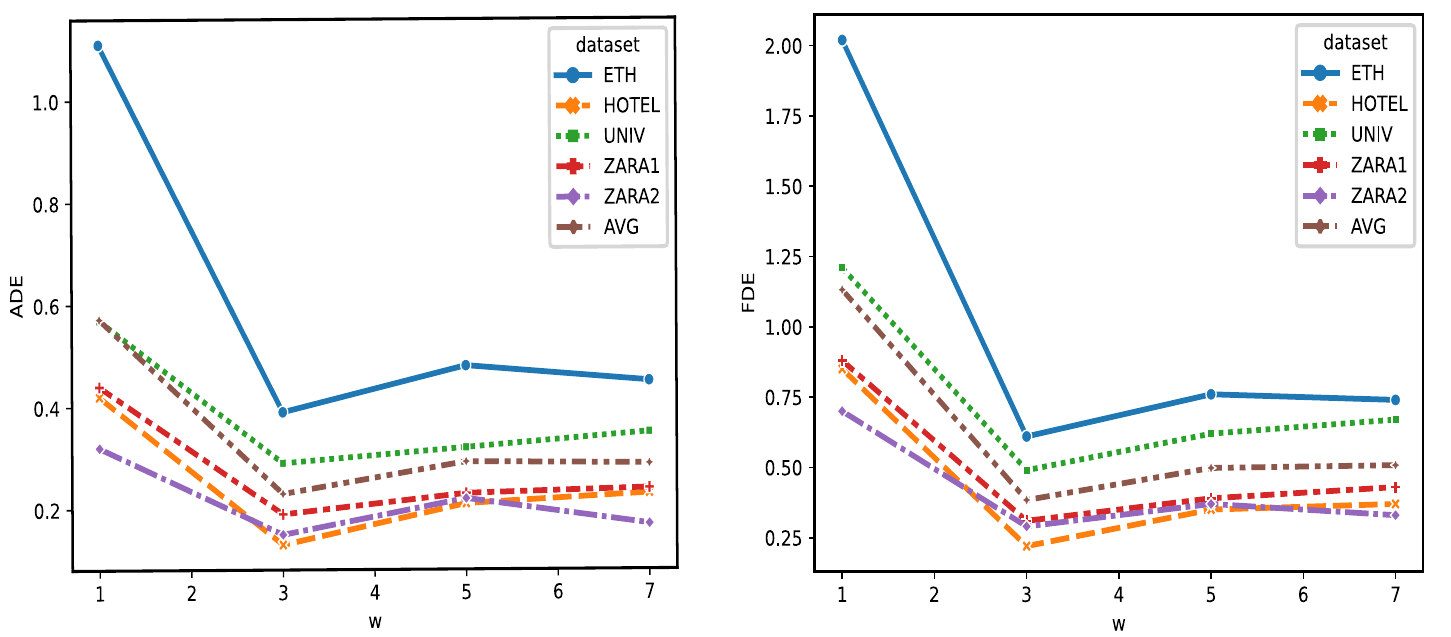} %
\caption{Illustration of the performance variation with different widths on ADE and FDE.}
\label{width_vis}
\end{figure}

\noindent\textbf{Performance on Different Widths of Tree}.
Similar to the depth, we also analyze the performance variation with different widths.
The coarse trajectory tree is set to split into $1, 3, 5, $ and $7$ directions, respectively.
The results are illustrated in Figure~\ref{width_vis}, where the tree also suffers from a width-performance dilemma because the deeper tree brings the difficult selection. 

\section{More Qualitative Analysis}

\noindent\textbf{Interpretability on ADE}.
Our promise in interpretability of qualitative analysis for the main paper that represents the interpretability of minimum ADE is shown in Table~\ref{interpretability_ade}, where it achieves significant performances~(96.58\% - 100.00\%) on standard best-of-20 prediction.
\begin{table}[ht!]
\centering
\begin{tabular}{c|ccccc}
\toprule
K           & ETH & HOTEL & UNIV & ZARA1 & ZARA2   \\
\midrule
1   & 38.12\% & 35.51\% & 9.81\% & 7.88\% & 9.41\% \\
5   & 53.03\% & 49.28\% & 29.46\% & 54.50\% & 42.61\% \\
10   & 69.61\% & 62.96\% & 54.55\% & 86.10\% & 60.43\% \\
15   & 92.26\% & 81.10\% & 80.42\% & 96.26\% & 94.18\% \\
20   & 100.00\% & 96.58\% & 96.81\% & 99.77\% & 99.86\% \\
\bottomrule
\end{tabular}
\caption{Top-$K$ accuracy of selecting the closest path of tree with minimum ADE on ETH-UCY.}
\label{interpretability_ade}
\end{table}

\noindent\textbf{Interpretability on SDD}.
Similar to the main paper, we conduct detailed statistics in the testing set of Stanford Drone to record the rate that selecting the closest path in different best-of-$K$ predictions.
The closest path is selected by the minimum FDE and minimum ADE with the coarse ground truth, respectively.
The experimental results are shown in Table~\ref{interpertability_sdd}, where the standard best-of-20 prediction brings 
the significant performances~(97.27\% and 99.01\%).
\begin{table}[ht!]
\centering
\scalebox{0.8}{
\begin{tabular}{c|ccccc}
\toprule
\diagbox{Metric}{K}           & 1 & 5 & 10 & 15 & 20   \\
\midrule
FDE & 38.60\% & 69.84\% & 85.64\% & 92.75\% & 97.27\% \\
ADE & 48.17\% & 68.22\% & 83.45\% & 94.48\% & 99.01\% \\
\bottomrule
\end{tabular}
}
\caption{Top-$K$ accuracy of selecting the closet path of tree on ETH-UCY~\cite{eth,ucy}.}
\label{interpertability_sdd}
\end{table}

\begin{figure*}[t]
\centering
\includegraphics[width=\textwidth]{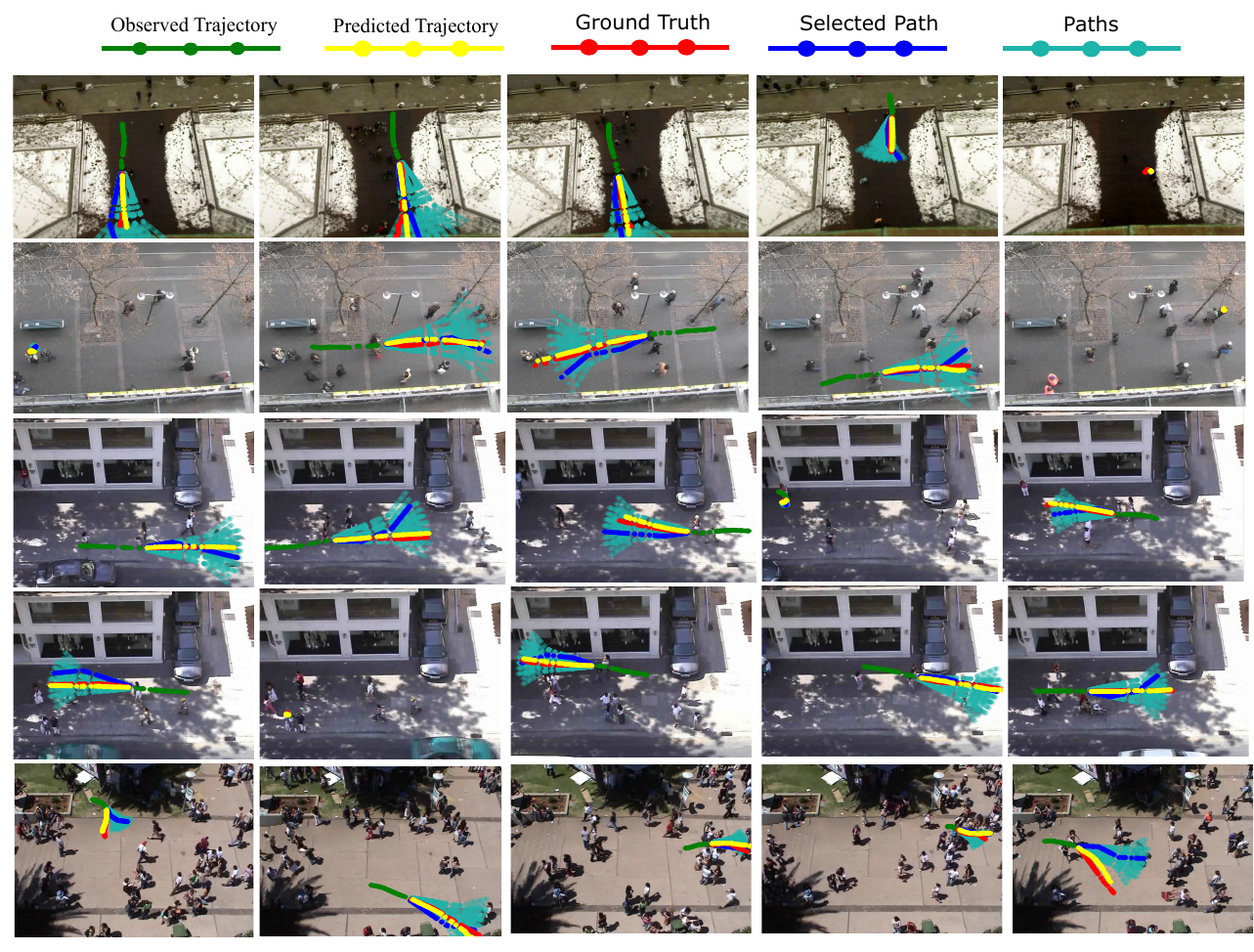} 
\caption{More visualizations of selected path and refined trajectory.}
\label{more_vis}
\end{figure*}

\noindent\textbf{More Visualizations}.
We show more visualizations randomly selected from various scenarios as illustrated in Figure~\ref{more_vis}.

\end{document}